\documentclass[runningheads]{llncs}
\usepackage{graphicx}
\usepackage{booktabs} 
\usepackage{amsmath,amssymb}
\usepackage{caption}
\usepackage{multirow}
\usepackage{adjustbox}
\usepackage{algorithmic}
\usepackage{algorithm}
\usepackage{bbm}
\usepackage{amsmath} 
\usepackage{color}
\usepackage{lmodern} 
\usepackage{diagbox}

\DeclareMathOperator*{\argmax}{arg\,max}
\DeclareMathOperator*{\argmin}{arg\,min}

\definecolor{todocolor}{rgb}{0.9,0.1,0.1}
\definecolor{lcolor}{rgb}{0.7,0.7,0.3}
\definecolor{qcolor}{rgb}{0,0,1}

\begin{document}
%

\title{CeFlow: A Robust and Efficient Counterfactual Explanation Framework for Tabular Data using Normalizing Flows}
\titlerunning{CeFlow: Counterfactual explanation with NF}

\author{Tri Dung Duong\inst{1} \and
Qian Li\inst{2} \and
Guandong Xu\inst{1}\thanks{Corresponding author: Guandong.Xu@uts.edu.au}}
%
%
\institute{Faculty of Engineering and Information Technology, University of Technology Sydney, NSW, Australia \and
School of Electrical Engineering, Computing and Mathematical Sciences,\\ Curtin University, WA, Australia
}
\maketitle              
\begin{abstract}
Counterfactual explanation is a form of interpretable machine learning that generates perturbations on a sample to achieve the desired outcome. The generated samples can act as instructions to guide end users on how to observe the desired results by altering samples. Although state-of-the-art counterfactual explanation methods are proposed to use variational autoencoder (VAE) to achieve promising improvements, they suffer from two major limitations: 1) the counterfactuals generation is prohibitively slow, which prevents algorithms from being deployed in interactive environments; 2) the counterfactual explanation algorithms produce unstable results due to the randomness in the sampling procedure of variational autoencoder. In this work, to address the above limitations, we design a robust and efficient counterfactual explanation framework, namely CeFlow, which utilizes normalizing flows for the mixed-type of continuous and categorical features. Numerical experiments demonstrate that our technique compares favorably to state-of-the-art methods. We release our source code\footnote{\url{https://github.com/tridungduong16/fairCE.git}} for reproducing the results.


\end{abstract}
\section{Introduction}
Machine learning (ML) has resulted in advancements in a variety of scientific and technical fields, including computer vision, natural language processing, and conversational assistants.
Interpretable machine learning is a machine learning sub-field that aims to provide a collection of tools, methodologies, and algorithms capable of producing high-quality explanations for machine learning model judgments. A great deal of methods in interpretable ML methods has been proposed in recent years. Among these approaches, counterfactual explanation (CE) is the prominent example-based method involved in how to alter features to change the model predictions and thus generates counterfactual samples for explaining and interpreting models \cite{mahajan2019preserving,artelt2020convex,grath_interpretable_2018,wachter2017counterfactual,xu2020causality}.
An example is that for a customer \texttt{A} rejected by a loan application, counterfactual explanation algorithms aim to generate counterfactual samples such as ``your loan would have been approved if your income was \$51,000 more'' which can act as a recommendation for a person to achieve the desired outcome. Providing counterfactual samples for black-box models has the capability to facilitate human-machine interaction, thus promoting the application of ML models in several fields. 

The recent studies in counterfactual explanation utilize variational autoencoder (VAE) as a generative model to generate counterfactual sample \cite{pawelczyk2020learning,mahajan2019preserving}. Specifically, the authors first build an encoder and decoder model from the training data. Thereafter, the original input would go through the encoder model to obtain the latent representation. They make the perturbation into this representation and pass the perturbed vector to the encoder until getting the desired output. However, these approaches present some limitations. First, the latent representation which is sampled from the encoder model would be changed corresponding to different sampling times, leading to unstable counterfactual samples. Thus, the counterfactual explanation algorithm is not robust when deployed in real applications. Second, the process of making perturbation into latent representation is so prohibitively slow \cite{mahajan2019preserving} since they need to add random vectors to the latent vector repeatedly; accordingly, the running time of algorithms grows significantly. Finally, the generated counterfactual samples are not closely connected to the density region, making generated explanations infeasible and non-actionable.  
To address all of these limitations, we propose a Flow-based counterfactual explanation framework (CeFlow) that integrates normalizing flow which is an invertible neural network as the generative model to generate counterfactual samples. Our contributions can be summarized as follows:
\begin{itemize}
    \item We introduce CeFlow, an efficient and robust counterfactual explanation framework that leverages the power of normalizing flows in modeling data distributions to generate counterfactual samples. The usage of flow-based models enables to produce more robust counterfactual samples and reduce the algorithm running time. 
    

    \item We construct a conditional normalizing flow model that can deal with tabular data consisting of continuous and categorical features by utilizing variational dequantization and Gaussian mixture models. 
    
    \item The generated samples from CeFlow are close to and related to high-density regions of other data points with the desired class. This makes counterfactual samples likely reachable and therefore naturally follow the distribution of the dataset. 
    
    

\end{itemize}




\section{Related works}
An increasing number of methods have been proposed for the counterfactual explanation. The existing methods can be categorized into gradient-based methods \cite{wachter2017counterfactual,mothilal2020explaining}, auto-encoder model \cite{mahajan2019preserving}, heuristic search methods \cite{poyiadzi2020face,sharma2020certifai} and integer linear optimization \cite{kanamori2020dace}. Regarding gradient-based methods, The authors in the study construct the cross-entropy loss between the desired class and counterfactual samples' prediction with the purpose of changing the model output. The created loss would then be minimized using gradient-descent optimization methods. In terms of auto-encoder model, generative models such as variational auto-encoder (VAE) is used to generate new samples in another line of research. The authors \cite{pawelczyk2020learning} first construct an encoder-decoder architecture. They then utilize the encoder to generate the latent representation, make some changes to it, and run it through the decoder until the prediction models achieve the goal class. However, VAE models which maximize the lower bound of the log-likelihood instead of measuring exact log-likelihood can produce unstable and unreliable results. On the other hand, there is an increasing number of counterfactual explanation methods based on heuristic search to select the best counterfactual samples such as Nelder-Mead \cite{grath2018interpretable}, growing spheres \cite{laugel2018comparison}, FISTA \cite{dhurandhar2019model,van2019interpretable}, or genetic algorithms \cite{dandl2020multi,lash2017generalized}. Finally, the studies \cite{ustun2019actionable} propose to formulate the problem of finding counterfactual samples as a mixed-integer linear optimization problem and utilize some existing solvers \cite{bliek1u2014solving,artelt2020convex} to obtain the optimal solution.

\section{Preliminaries}

Throughout the paper, lower-cased letters $x$ and $\boldsymbol{x}$ denote the deterministic scalars and vectors, respectively. We consider a classifier $\mathcal{H}: \mathcal{X} \rightarrow \mathcal{Y}$ that has the input of feature space $\mathcal{X}$ and the output as $\mathcal{Y} = \{1 ... \mathcal{C}\}$ with $\mathcal{C}$ classes. Meanwhile, we denote a dataset $\mathcal{D} = \{\boldsymbol{x}_n, y_n\}^N_{n=1}$ consisting of $N$ instances where $\boldsymbol{x}_n \in \mathcal{X}$ is a sample, $y_n \in \mathcal{Y}$ is the predicted label of individuals $\boldsymbol{x}_n$ from the classifier $\mathcal{H}$. Moreover, $f_\theta$ is denoted for a normalizing flow model parameterized by $\theta$. Finally, we split the feature space into two disjoint feature subspaces of categorical features and continuous features represented by $\mathcal{X}^\text{cat}$ and $\mathcal{X}^\text{con}$ respectively such that $\mathcal{X} = \mathcal{X}_\text{cat} \times \mathcal{X}_\text{con}$ and $\boldsymbol{x}=(\boldsymbol{x}^\text{cat}, \boldsymbol{x}^\text{con})$, and $\boldsymbol{x}^{\text{cat}_j}$ and $\boldsymbol{x}^{\text{con}_j}$ is
the corresponding $j$-th feature of $\boldsymbol{x}^\text{cat}$ and $\boldsymbol{x}^\text{con}$.


\subsection{Counterfactual explanation}
\label{cf:objective}

With the original sample $\boldsymbol{x}_\text{org} \in \mathcal{X}$ and its predicted output $y_\text{org} \in \mathcal{Y}$, the counterfactual explanation aims to find the nearest counterfactual sample $\boldsymbol{x}_\text{cf}$ such that the outcome of classifier for $\boldsymbol{x}_\text{cf}$ is changed to desired output class $y_\text{cf}$. We aim to identify the perturbation $\boldsymbol{\delta}$ such that counterfactual instance $\boldsymbol{x}_\text{cf} = \boldsymbol{x}_\text{org} + \boldsymbol{\delta}$ is the solution of the following optimization problem:


\begin{equation}
\label{eqn:original}
\small 
\boldsymbol{x}_\text{cf} = \argmin_{\boldsymbol{x}_\text{cf} \in \mathcal{X}} d(\boldsymbol{x}_\text{cf}, \boldsymbol{x}_\text{org}) \quad\text{subject to}\quad \mathcal{H}(\boldsymbol{x}_\text{cf}) = y_\text{cf}
\end{equation}
where $d(\boldsymbol{x}_\text{cf}, \boldsymbol{x}_\text{org})$ is the function measuring the distance between $\boldsymbol{x}_\text{org}$ and $\boldsymbol{x}_\text{cf}$. Eq~\eqref{eqn:original} demonstrates the optimization objective that minimizes the similarity of the counterfactual and original samples, as well as ensures to change the classifier to the desirable outputs. To make the counterfactual explanations plausible, they should only suggest minimal changes in features of the original sample. \cite{mothilal2020explaining}.

\subsection{Normalizing flow}
\label{nf:objective}

Normalizing flows (NF) \cite{dinh2014nice} is the active research direction in generative models that aims at modeling the probability distribution of a given dataset. The study \cite{dinh2016density} first proposes a normalizing flow, which is an unsupervised density estimation model described as an invertible mapping $f_\theta: \mathcal{X} \rightarrow \mathcal{Z}$ from the data space $\mathcal{X}$ to the latent space $\mathcal{Z}$. Function $f_\theta$ can be designed as a neural network parametrized by $\theta$ with architecture that has to ensure invertibility and efficient computation of log-determinants. The data distribution is modeled as a transformation $f_\theta^{-1}: \mathcal{Z} \rightarrow \mathcal{X}$ applied to a random variable from the latent distribution $\boldsymbol{z} \sim p_{\mathcal{Z}}$, for which Gaussian distribution is chosen. The change of variables formula gives the density of the converted random variable $\boldsymbol{x}=f_\theta^{-1}(\boldsymbol{z})$ as follows:
\begin{equation}
\label{normalizingflowobjective0}
\small
\begin{aligned}
    p_{\mathcal{X}}(\boldsymbol{x}) & =p_{\mathcal{Z}}(f_\theta(\boldsymbol{x})) \cdot\left|\operatorname{det}\left(\frac{\partial f_\theta}{\partial \boldsymbol{x}}\right)\right|\\
         & \propto \log \left(p_{\mathcal{Z}}(f_\theta(\boldsymbol{x}))\right)  + \log \left(\left|\operatorname{det}\left(\frac{\partial f_\theta}{\partial \boldsymbol{x}}\right)\right|\right)
\end{aligned}
\end{equation}
With $N$ training data points $\mathcal{D} = \{\boldsymbol{x}_n\}_{n=1}^N$, the model with respects to parameters $\theta$ can be trained by maximizing the likelihood in Equation \eqref{loglikelihood}:
\begin{equation}
\small 
\label{loglikelihood}
   \theta = \argmax_{\theta} \left( \prod_{n=1}^N \left(\log (p_{\mathcal{Z}}(f_\theta(\boldsymbol{x}_n))) + \log \left(\left|\operatorname{det}\left(\frac{\partial f_\theta(\boldsymbol{x}_n)}{\partial \boldsymbol{x}_n}\right)\right|\right) \right) \right)
\end{equation}

\section{Methodology}
In this section, we illustrate our approach (CeFlow) which leverages the power of normalizing flow in generating counterfactuals. First, we define the general architecture of our framework in section~\ref{ps}. Thereafter, section~\ref{nfcat} and \ref{nfarch} illustrate how to train and build the architecture of the invertible function $f$ for tabular data, while section~\ref{nfperturb} describes how to produce the counterfactual samples by adding the perturbed vector into the latent representation. 

\subsection{General architecture of CeFlow}
\label{ps}

\begin{figure*}[!htb]
\centering
\includegraphics[width=1.0\textwidth]{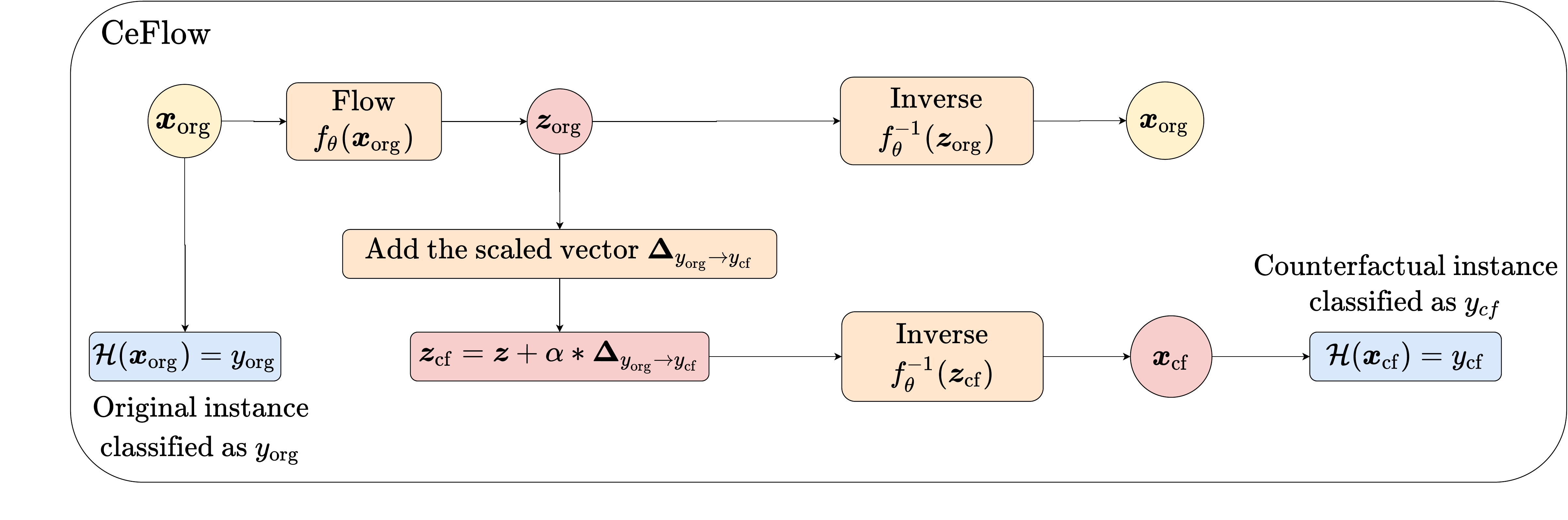}
\caption{Counterfactual explanation with normalizing flows (CeFlow).}
\label{fig:generalachitecture}
\end{figure*}

Figure~\ref{fig:generalachitecture} generally illustrates our framework. Let $\boldsymbol{x}_\text{org}$ be an original instance, and ${f}_\theta$ denote a pre-trained, invertible and differentiable normalizing flow model on the training data. In general, we first construct an invertible and differentiable function $f_\theta$ that converts the original instance $\boldsymbol{x}_\text{org}$ to the latent representation $\boldsymbol{z}_\text{org} = f(\boldsymbol{x}_\text{org})$. After that, we would find the scaled vector $\boldsymbol{\delta}_{z}$ as the perturbation and add to the latent representation $\boldsymbol{z}_\text{org}$ to get the perturbed representation ${\boldsymbol{z}}_\text{cf}$ which goes through the inverse function $f_\theta^{-1}$ to produce the counterfactual instance $\boldsymbol{x}_\text{cf}$. With the counterfactual instance $\boldsymbol{x}_\text{cf} = f_\theta^{-1}(\boldsymbol{z}_\text{org} + \boldsymbol{\delta}_{z})$, we can re-write the objective function Eq.~\eqref{eqn:original} into the following form:

\begin{equation}
\label{eqn:distance}
\small 
\begin{cases}
\boldsymbol{\delta}_{z} = \argmin_{\boldsymbol{\delta}_{z} \in \mathcal{Z}} d(\boldsymbol{x}_\text{org}, \boldsymbol{\delta}_{z}) \\ 
        \mathcal{H}(\boldsymbol{x}_\text{cf}) = y_\text{cf} 
\end{cases}
\end{equation}





One of the biggest problems of deploying normalizing flow is how to handle mixed-type data which contains both continuous and categorical features. Categorical features are in discrete forms, which is challenging to model by the continuous distribution only \cite{ho2019flow++}. Another challenge is to construct the objective function to learn the conditional distribution on the predicted labels
\cite{winkler2019learning,izmailov2020semi}. In the next section, we will discuss how to construct the conditional normalizing flow $f_\theta$ for tabular data. 





\subsection{Normalizing flows for categorical features}
\label{nfcat}
This section would discuss how to handle the categorical features. Let $\{\boldsymbol{z}^{\text{cat}_m}\}_{m=1}^M$ be the continuous representation of $M$ categorical features $\{\boldsymbol{x}^{\text{cat}_m}\}_{m=1}^M$ for each $\boldsymbol{x}^{\text{cat}_m} \in \{0,1,...,K-1\}$ with $K > 1$.  Follow by several studies in the literature \cite{ho2019flow++,hoogeboom2020learning}, we utilize variational dequantization to model the categorical features. The key idea of variational dequantization is to add noise $\boldsymbol{u}$ to the discrete values $\boldsymbol{x}^{\text{cat}}$ to convert the discrete distribution $p_{\mathcal{X}^{\text{cat}}}$ into a continuous distribution $p_{\phi_\text{cat}}$. With $\boldsymbol{z}^{\text{cat}} = \boldsymbol{x}^{\text{cat}}+\boldsymbol{u}_k$, $\phi_\text{cat}$ and $\theta_\text{cat}$ be models' parameters, we have following objective functions:
\begin{equation} 
\small 
\label{loglikelihoodcate1}
\begin{split}
\log p_{\mathcal{X}^{\text{cat}}}(\boldsymbol{x}^{\text{cat}}) 
&\geq \int_{\boldsymbol{u}} \log \frac{p_{\phi_\text{cat}}(\boldsymbol{z}^{\text{cat}})}{q_{\theta_\text{cat}}(\boldsymbol{u}|\boldsymbol{x}^{\text{cat}})}du \\
& \approx \frac{1}{K} \sum_{k=1}^K \log \prod_{m=1}^M \frac{p_{\phi_\text{cat}}(\boldsymbol{x}^{\text{cat}_m} + \boldsymbol{u}_k)}{q_{\theta_\text{cat}}(\boldsymbol{u}_k|\boldsymbol{x}^{\text{cat}})} 
\end{split}
\end{equation}

Followed the study \cite{hoogeboom2020learning}, we choose Gaussian dequantization which is more powerful than the uniform dequantization as $q_{\theta_\text{cat}}(\boldsymbol{u}_k|\boldsymbol{x}^{\text{cat}}) = \text{sig} \left( \mathcal{N}\left(\boldsymbol{\mu}_{\theta_\text{cat}}, \boldsymbol{\Sigma}_{\theta_\text{cat}}\right)\right)$ with mean $\boldsymbol{\mu}_{\theta_\text{cat}}$, covariance $\boldsymbol{\Sigma}_{\theta_\text{cat}}$ and sigmoid function sig(·).



\subsection{Conditional Flow Gaussian Mixture Model for tabular data}
\label{nfarch}

The categorical features $\boldsymbol{x}^{\text{cat}}$ going through the variational dequantization would convert into continuous representation $\boldsymbol{z}^{\text{cat}}$. We then perform merge operation on continuous representation $\boldsymbol{z}^\text{cat}$ and continuous feature $\boldsymbol{x}^\text{con}$ to obtain values $\left(\boldsymbol{z}^\text{cat}, \boldsymbol{x}^\text{con}\right) \mapsto \boldsymbol{x}^\text{full}$. Thereafter, we apply flow Gaussian mixture model \cite{izmailov2020semi} which is a probabilistic generative model for training the invertible function $f_\theta$. For each predicted class label $y \in\{1. . .\mathcal{C}\}$, the latent space distribution $p_{\mathcal{Z}}$ conditioned on a label $k$ is the Gaussian distribution $\mathcal{N}\left(\boldsymbol{z}^\text{full} \mid \boldsymbol{\mu}_{k}, \boldsymbol{\Sigma}_{k}\right)$ with mean $\boldsymbol{\mu}_{k}$ and covariance $\boldsymbol{\Sigma}_{k}$:
\begin{equation}
\small
p_{\mathcal{Z}}(\boldsymbol{z}^\text{full} \mid y=k)=\mathcal{N}\left(\boldsymbol{z}^\text{full} \mid \boldsymbol{\mu}_{k}, \boldsymbol{\Sigma}_{k}\right)
\end{equation}
As a result, we can have the marginal distribution of $\boldsymbol{z}^\text{full}$:
\begin{equation}
\label{marginaldistributionz}
\small 
p_{\mathcal{Z}}(\boldsymbol{z}^\text{full})=\frac{1}{\mathcal{C}} \sum_{k=1}^{\mathcal{C}} \mathcal{N}\left(\boldsymbol{z}^\text{full} \mid \boldsymbol{\mu}_{k}, \boldsymbol{\Sigma}_{k}\right)
\end{equation}
The density of the transformed random variable $\boldsymbol{x}^\text{full} = f_\theta^{-1}(\boldsymbol{z}^\text{full})$ is given by:
\begin{equation}
\small
\label{normalizingflowobjective}
    p_{\mathcal{X}}(\boldsymbol{x}^\text{full}) = \log \left(p_{\mathcal{Z}}(f_\theta(\boldsymbol{x}^\text{full}))\right)  + \log \left(\left|\operatorname{det}\left(\frac{\partial f_\theta}{\partial \boldsymbol{x}^\text{full}}\right)\right|\right)
\end{equation}
Eq.~\eqref{marginaldistributionz} and Eq.~\eqref{normalizingflowobjective} together lead to the likelihood for data as follows:
\begin{equation}
\label{loglikelihoodcontinuous}
\small 
p_{\mathcal{X}}(\boldsymbol{x}^\text{full} \mid y=k)=\log \left(\mathcal{N}\left(f_\theta(\boldsymbol{x}^\text{full}) \mid \boldsymbol{\mu}_{k}, \boldsymbol{\Sigma}_{k}\right)\right) + \log \left(\left|\operatorname{det}\left(\frac{\partial f_\theta}{\partial \boldsymbol{x}^\text{full}}\right)\right|\right)
\end{equation}
We can train the model by maximizing the joint likelihood of the categorical and continuous features on $N$ training data points $\mathcal{D} = \{(\boldsymbol{x}_n^{\text{con}}, \boldsymbol{x}_n^{\text{cat}})\}_{n=1}^N$ by combining Eq.~\eqref{loglikelihoodcate1} and Eq.~\eqref{loglikelihoodcontinuous}:
\begin{equation}
\label{normalizingflowobjective0}
\small 
\begin{aligned}
 \theta^*, \phi_\text{cat}^*, \theta_\text{cat}^* &= \argmax_{\theta, \phi_\text{cat}, \theta_\text{cat}} \prod_{n=1}^{N} \left( \prod_{\boldsymbol{x}_n^{\text{con}} \in \mathcal{X}^{\text{con}}} p_{\mathcal{X}}\left(\boldsymbol{x}_n^{\text{con}}\right) \prod_{\boldsymbol{x}_n^{\text{cat}} \in \mathcal{X}^{\text{cat}}} p_{\mathcal{X}}\left(\boldsymbol{x}_n^{\text{cat}}\right) \right) \\
 &= \argmax_{\theta, \phi_\text{cat}, \theta_\text{cat}} \prod_{n=1}^{N} \left( \log \left(\mathcal{N}\left(f_\theta(\boldsymbol{x}_n^\text{full}) \mid \boldsymbol{\mu}_{k}, \boldsymbol{\Sigma}_{k}\right)\right) + \log \left(\left|\operatorname{det}\left(\frac{\partial f_\theta}{\partial \boldsymbol{x}_n^\text{full}}\right)\right|\right) \right)
\end{aligned}
\end{equation}
\subsection{Counterfactual generation step}
\label{nfperturb}

In order to find counterfactual samples, the recent approaches \cite{mothilal2020explaining,wachter2017counterfactual} normally define the loss function and deploy some optimization algorithm such as gradient descent or heuristic search to find the perturbation. These approaches however demonstrates the prohibitively slow running time, which prevents from deploying in interactive environment\cite{holtgen2021deduce}. Therefore, inspired by the study \cite{hvilshoj2021ecinn}, we add the scaled vector as the perturbation from the original instance $\boldsymbol{x}_\text{org}$ to counterfactual one $\boldsymbol{x}_\text{cf}$. By Bayes’ rule, we notice that under a uniform prior distribution over labels $p(y=k) = \frac{1}{\mathcal{C}}$ for $\mathcal{C}$ classes, the log posterior probability becomes:
\begin{equation}
\small 
\label{logequation}
    \log p_\mathcal{X}(y=k|\boldsymbol{x})=\log\frac{p_\mathcal{X}(\boldsymbol{x}|y=k)}{\sum_{k=1}^{\mathcal{C}}p_\mathcal{X}(\boldsymbol{x}|y=k)} \propto \left| \left| f_{\theta}(\boldsymbol{x}) - \boldsymbol{\mu}_k\right|\right|^2
\end{equation}
We observed from Eq.~\eqref{logequation} that latent vector $\boldsymbol{z} = f_{\theta}(\boldsymbol{x})$ will be predicted from the class $y$ with the closest model mean $\boldsymbol{\mu}_k$. For each predicted class $k \in \{1...\mathcal{C}\}$, we denote $\mathcal{G}_k = \{ {\boldsymbol{x}_m, y_m}\}_{m=1}^M$ as a set of $M$ instances with the same predicted class as $y_m=k$. We define the mean latent vector $\boldsymbol{\mu}_k$ corresponding to each class $k$ such that:
\begin{equation}
\small
    \boldsymbol{\mu}_k = \frac{1}{M}\sum_{\boldsymbol{x}_m \in \mathcal{G}_k}{f_{\theta}(\boldsymbol{x}_m)}
    \label{meanvector}
\end{equation}
Therefore, the scaled vector that moves the latent vector $\boldsymbol{z}_\text{org}$ to the decision boundary from the original class $y_\text{org}$ to counterfactual class $y_\text{cf}$ is defined as:
\begin{equation}
\small
    \boldsymbol{\Delta}_{y_\text{org} \rightarrow y_\text{cf}} = \left| \boldsymbol{\mu}_{y_\text{org}}- \boldsymbol{\mu}_{y_\text{cf}} \right|
\end{equation}

The scaled vector $\boldsymbol{\Delta}_{y_\text{org} \rightarrow y_\text{cf}}$ is added to the original latent representation $\boldsymbol{z}_\text{cf} = f_\theta(\boldsymbol{x}_\text{org})$ to obtained the perturbed vector. The perturbed vector then goes through inverted function $f_\theta^{-1}$ to re-produce the counterfactual sample:


\begin{equation}
\small 
\label{counterfactual}
    \boldsymbol{x}_\text{cf} = f_{\theta}^{-1}(f_{\theta}(\boldsymbol{x}_\text{org}) + \alpha\boldsymbol{\Delta}_{y_\text{org} \rightarrow y_\text{cf}})
\end{equation}

We note that the hyperparameter $\alpha$ needs to be optimized by searching in a range of values. The full algorithm is illustrated in Algorithm~\ref{alg:mulobj}.

    \scalebox{0.96}{
    \begin{minipage}{\linewidth}
\begin{algorithm}[H]
\small
\caption{Counterfactual explanation flow (CeFlow)}
\label{alg:mulobj}
\begin{algorithmic}[1]
\renewcommand{\algorithmicrequire}{\textbf{Input:}}
\renewcommand{\algorithmicensure}{\textbf{Output:}}
\REQUIRE An original sample $\boldsymbol{x}_\text{org}$ with its prediction $y_\text{org}$, desired class $y_\text{cf}$, a provided machine learning classifier $\mathcal{H}$ and encoder model $Q_{\phi}$.
\STATE Train the invertible function $f_\theta$ by maximizing the log-likelihood: 
$$
\begin{aligned}
 \theta^*, \phi_\text{cat}^*, \theta_\text{cat}^* &= \argmax_{\theta, \phi_\text{cat}, \theta_\text{cat}} \prod_{n=1}^{N} \left( \prod_{\boldsymbol{x}_n^{\text{con}} \in \mathcal{X}^{\text{con}}} p_{\mathcal{X}}\left(\boldsymbol{x}_n^{\text{con}}\right) \prod_{\boldsymbol{x}_n^{\text{cat}} \in \mathcal{X}^{\text{cat}}} p_{\mathcal{X}}\left(\boldsymbol{x}_n^{\text{cat}}\right) \right) \\
 &= \argmax_{\theta, \phi_\text{cat}, \theta_\text{cat}} \prod_{n=1}^{N} \left( \log \left(\mathcal{N}\left(f_\theta(\boldsymbol{x}_n^\text{full}) \mid \boldsymbol{\mu}_{k}, \boldsymbol{\Sigma}_{k}\right)\right) + \log \left(\left|\operatorname{det}\left(\frac{\partial f_\theta}{\partial \boldsymbol{x}_n^\text{full}}\right)\right|\right) \right)
\end{aligned}
$$
\STATE Compute mean latent vector $\boldsymbol{\mu}_{k}$ for each class $k$ by $\boldsymbol{\mu}_k = \frac{1}{M}\sum_{\boldsymbol{x}_m \in \mathcal{G}_k}{f(\boldsymbol{x}_m)}$.
\STATE Compute the scaled vector $\boldsymbol{\Delta}_{y_\text{org} \rightarrow y_\text{cf}} = \left| \boldsymbol{\mu}_{y_\text{org}}- \boldsymbol{\mu}_{y_\text{cf}} \right|$.
\STATE Find the optimal hyperparameter $\alpha$ by searching a range of values. 
\STATE Compute $    \boldsymbol{x}_\text{cf} = f_{\theta}^{-1}(f_{\theta}(\boldsymbol{x}_\text{org}) + \alpha\boldsymbol{\Delta}_{y_\text{org} \rightarrow y_\text{cf}})
$.
\ENSURE $\boldsymbol{x}_\text{cf}$.
\end{algorithmic}
\end{algorithm}
\end{minipage}%
}

\section{Experiments}

We run experiments on three datasets to show that our method outperforms state-of-the-art approaches. The specification of hardware for the experiment is Python 3.8.5 with 64-bit Red Hat, Intel(R) Xeon(R) Gold 6238R CPU @ 2.20GHz. We implement our algorithm by using Pytorch library and adopt the RealNVP architecture \cite{dinh2016density}. During training progress, Gaussian mixture parameters are fixed: the means are initialized randomly from the standard normal distribution and the covariances are
set to ${I}$. More details of implementation settings can be found in our code repository\footnote{\url{https://anonymous.4open.science/r/fairCE-538B}}.  

We evaluate our approach via three datasets: \texttt{Law} \cite{wightman1998lsac}, \texttt{Compas} \cite{larson2016we} and \texttt{Adult} \cite{Dua:2019}. \texttt{Law}\footnote{\url{http://www.seaphe.org/databases.php}}\cite{wightman1998lsac} dataset provides information of students with their features: their entrance exam scores (LSAT), grade-point average (GPA) and first-year average grade (FYA). \texttt{Compas}\footnote{\url{https://www.propublica.org}}\cite{larson2016we} dataset contains information about 6,167 prisoners who have features including gender, race and other attributes related to prior conviction and age. \texttt{Adult}\footnote{\url{https://archive.ics.uci.edu/ml/datasets/adult}}\cite{Dua:2019} dataset is a real-world dataset consisting of both continuous and categorical features of a group of consumers who apply for a loan at a financial institution. 

We compare our proposed method (CeFlow) with several state-to-the-art methods including \text{Actionable Recourse (AR)} \cite{ustun2019actionable}, \text{Growing Sphere (GS)} \cite{laugel2017inverse}, \text{FACE} \cite{poyiadzi2020face}, \text{CERTIFAI} \cite{sharma2020certifai}, DiCE \cite{mothilal2020explaining} and \text{C-CHVAE} \cite{pawelczyk2020learning}. Particularly, we implement the CERTIFAI with library PyGAD\footnote{\url{https://github.com/ahmedfgad/GeneticAlgorithmPython}} and utilize the available source code\footnote{\url{https://github.com/divyat09/cf-feasibility}} for implementation of DiCE, while other approaches are implemented with Carla library \cite{pawelczyk2021carla}. Finally, we report the results of our proposed model on a variety of metrics 
including \text{success rate} (success), \text{${l}_1$-norm} ($l_1$), \text{categorical proximity} \cite{mothilal2020explaining}, \text{continuous proximity} \cite{mothilal2020explaining} and \text{mean log-density} \cite{artelt2020convex}. Note that for $l_1$-norm, we report mean and variance of \text{$l_1$-norm} corresponding to \text{$l_1$-mean} and \text{$l_1$-variance}. Lower \text{$l_1$-variance} aims to illustrate the algorithm's robustness.

\begin{table*}[!htb]
\caption{Performance of all methods on the classifier.
We compute $p$-value by conducting a paired $t$-test between our approach (CeFlow) and baselines with 100 repeated experiments for each metric. }
\begin{adjustbox}{width=0.7\columnwidth,center}
\begin{scriptsize}

\begin{tabular}{@{}lcccccccc@{}}
\toprule
\multirow{2}{*}{\textbf{Dataset}} & \multirow{2}{*}{\textbf{Method}} & \multicolumn{4}{c}{\textbf{Performance}}                                                                           & \multicolumn{3}{c}{\textbf{p-value}}                            \\ \cmidrule(l){3-6} \cmidrule(l){7-9}
    &                           & \textbf{success  } & \textbf{$l_1$-mean  } & \multicolumn{1}{c}{\textbf{$l_1$-var  }} & \textbf{log-density  } & \textbf{success} & \textbf{$l_1$} & \textbf{log-density }  \\ 
    \midrule
\multirow{7}{*}{\texttt{Law}}        & AR                               & 98.00                                  & 3.518                                  & 2.0e-03                           & -0.730                                  & 0.041                             & 0.020                           & 0.022                                  \\
                                  & GS                               & 100.00                              & 3.600                                  & 2.6e-03                           & -0.716                                  & 0.025                             & 0.048                           & 0.016                                  \\
                                  & FACE                             & 100.00                              & 3.435                                  & 2.0e-03                           & -0.701                                  & 0.029                             & 0.010                           & 0.017                                  \\
                                  & CERTIFAI                         & 100.00                              & 3.541                                  & 2.0e-03                           & -0.689                                  & 0.029                             & 0.017                           & 0.036                                  \\
                                  & DiCE                             & 94.00                               & \textbf{3.111}        & 2.0e-03                           & -0.721                                  & 0.018                             & 0.035                           & 0.048                                  \\
                                  & C-CHVAE                          & 100.00                               & 3.461                                  & 1.0e-03                           & -0.730                                  & 0.040                             & 0.037                           & 0.016                                  \\
                                  & CeFlow                           & \textbf{100.00}    & \text{3.228}          & \textbf{1.0e-05} & -\textbf{0.679}        & -                                 & -                               & -                                      \\ \midrule
\multirow{7}{*}{\texttt{Compas}}           & AR                               & 97.50                                  & 1.799                                  & 2.4e-03                           & -14.92                                  & 0.038                             & 0.034                           & 0.046                                  \\
                                  & GS                               & 100.00                              & 1.914                                  & 3.2e-03                           & -14.87                                  & 0.019                             & 0.043                           & 0.040                                  \\
                                  & FACE                             & 98.50                                & 1.800                                  & 4.8e-03                           & -15.59                                  & 0.036                             & 0.024                           & 0.035                                  \\
                                  & CERTIFAI                         & 100.00                              & 1.811                                  & 2.4e-03                           & -15.65                                  & 0.040                             & 0.048                           & 0.038                                  \\
                                  & DiCE                             & 95.50                               & 1.853                                  & 2.9e-03                           & -14.68                                  & 0.030                             & 0.029                           & 0.018                                  \\
                                  & C-CHVAE                          & 100.00                               & 1.878                                  & 1.1e-03                           & -13.97                                 & 0.026                             & 0.015                           & 0.027                                  \\
                                  & CeFlow                           & \textbf{100.00}    & \textbf{1.787}        & \textbf{1.8e-05} & -\textbf{13.62}        & -                                 & -                               & -                                      \\ \midrule
\multirow{7}{*}{\texttt{Adult}}           & AR                               & 100.00                               & 3.101                                  & 7.8e-03                           & -25.68                                  & 0.044                             & 0.037                           & 0.018                                  \\
                                  & GS                               & 100.00                              & 3.021                                  & 2.4e-03                           & -26.55                                  & 0.026                             & 0.049                           & 0.028                                  \\
                                  & FACE                             & 100.00                              & 2.991                                  & 6.6e-03                           & -23.57                                  & 0.027                             & 0.015                           & 0.028                                  \\
                                  & CERTIFAI                         & 93.00                               & 3.001                                  & 4.1e-03                           & -25.55                                  & 0.028                             & 0.022                           & 0.016                                  \\
                                  & DiCE                             & 96.00                               & 2.999                                  & 9.1e-03                           & -24.33                                  & 0.046                             & 0.045                           & 0.045                                  \\
                                  & C-CHVAE                          & 100.00                               & 3.001                                  & 8.7e-03                           & -24.45                                  & 0.026                             & 0.043                           & 0.019                                  \\
                                  & CeFlow                           & \textbf{100.00}    & \textbf{2.964}        & \textbf{1.5e-05} & -\textbf{23.46}        & -                             & -                           & -                                  \\ \bottomrule
\end{tabular}
\end{scriptsize}
\end{adjustbox}
\label{tab:result}
\end{table*}

\begin{table*}[!htb]
\caption{We report running time of different methods on three datasets.}
\begin{adjustbox}{width=1\columnwidth,center}
\begin{scriptsize}
\begin{tabular}{@{}rrrrrrrr@{}}
\toprule
\multicolumn{1}{l}{\textbf{Dataset}} & \multicolumn{1}{c}{\textbf{AR}}                    & \multicolumn{1}{c}{\textbf{GS}}                    & \multicolumn{1}{c}{\textbf{FACE}}                   & \multicolumn{1}{c}{\textbf{CERTIFAI}}                & \multicolumn{1}{c}{\textbf{DiCE}}                    & \multicolumn{1}{c}{\textbf{C-CHVAE}}                 & \multicolumn{1}{c}{\textbf{CeFlow}}                                          \\ \midrule
\multicolumn{1}{l}{\textbf{Law}}                           & 3.030  $\pm$    0.105 & 7.126    $\pm$  0.153 & 6.213    $\pm$  0.007  & 6.522    $\pm$  0.088  & 8.022    $\pm$  0.014  & 9.022    $\pm$  0.066  & \textbf{0.850 $\pm$  0.055} \\
\multicolumn{1}{l}{\textbf{Compas}}                                 & 5.125    $\pm$  0.097 & 8.048    $\pm$  0.176 & 7.688    $\pm$  0.131  & 13.426    $\pm$  0.158 & 7.810    $\pm$  0.076  & 6.879    $\pm$  0.044 & \textbf{0.809    $\pm$  0.162}   \\
\multicolumn{1}{l}{\textbf{Adult}}                                 & 7.046    $\pm$  0.151 & 6.472    $\pm$  0.021 & 13.851    $\pm$  0.001 & 7.943    $\pm$  0.046  & 11.821    $\pm$  0.162 & 12.132    $\pm$  0.024 & \textbf{0.837    $\pm$  0.026}  \\ \bottomrule
\end{tabular}
\end{scriptsize}
\end{adjustbox}
\label{tab:runningtime}
\end{table*}


\begin{figure}[!ht]
\centerline{\includegraphics[width=0.8\textwidth]{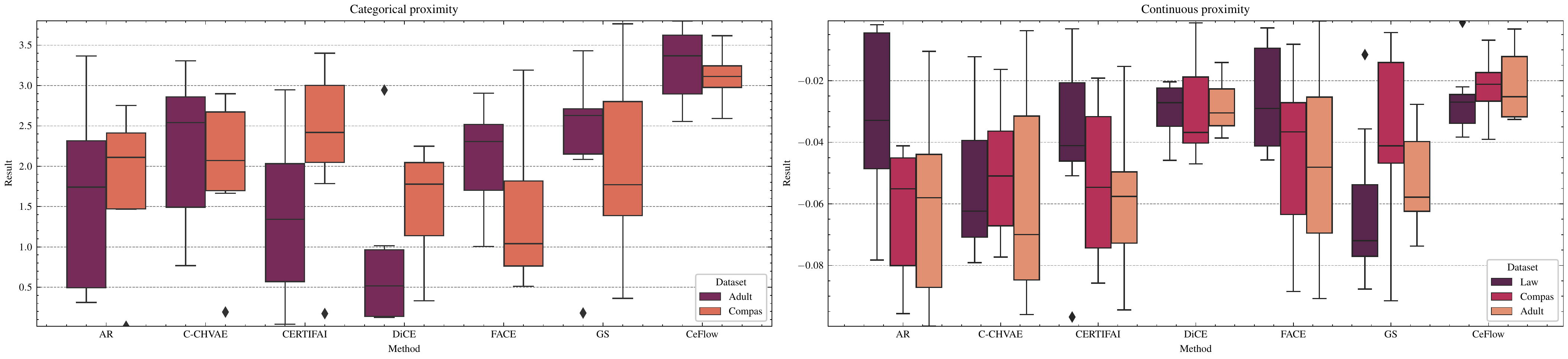}}
\caption{Baseline results in terms of \textbf{Categorical proximity} and \textbf{Continuous proximity}. Higher continuous and categorical proximity are better.}
\label{fig:proximity}
\end{figure}

The performance of different approaches regarding three metrics: $l_1$, success metrics and log-density are illustrated in Table~\ref{tab:result}. Regarding success rate, all three methods achieve competitive results, except the AR, DiCE and CERTIFAI performance in all datasets with around 90\% of samples belonging to the target class. These results indicate that by integrating normalizing flows into counterfactuals generation, our proposed method can achieve the target of counterfactual explanation task for changing the models' decision. Apart from that, for $l_1$-mean, CeFlow is ranked second with 3.228 for \texttt{Law}, and is ranked first for \texttt{Compas} and \texttt{Adult} (1.787 and 2.964). Moreover, our proposed method generally achieves the best performance regarding $l_1$-variance on three datasets. CeFlow also demonstrates the lowest log-density metric in comparison with other approaches achieving at -0.679, -13.62 and -23.46 corresponding to \texttt{Law}, \texttt{Compas} and \texttt{Adult} dataset. This illustrates that the generated samples are more closely followed the distribution of data than other approaches. We furthermore perform a statistical significance test to gain more insights into the effectiveness of our proposed method in producing counterfactual samples compared with other approaches. Particularly, we conduct the paired $t$-test between our approach (CeFlow) and other methods on each dataset and each metric with the obtained results on 100 randomly repeated experiments and report the result of $p$-value in Table~\ref{tab:result}. We discover that our model is statistically significant with $p < 0.05$, proving CeFlow's effectiveness in counterfactual samples generation tasks. Meanwhile, Table~\ref{tab:runningtime} shows the running time of different approaches. Our approach achieves outstanding performance with the running time demonstrating around 90\% reduction compared with other approaches. Finally, as expected, by using normalizing flows, CeFlow produces more robust counterfactual samples with the lowest $l_1$-variance and demonstrates an effective running time in comparison with other approaches. 

\begin{figure}[!ht]
\centerline{\includegraphics[width=0.6\textwidth]{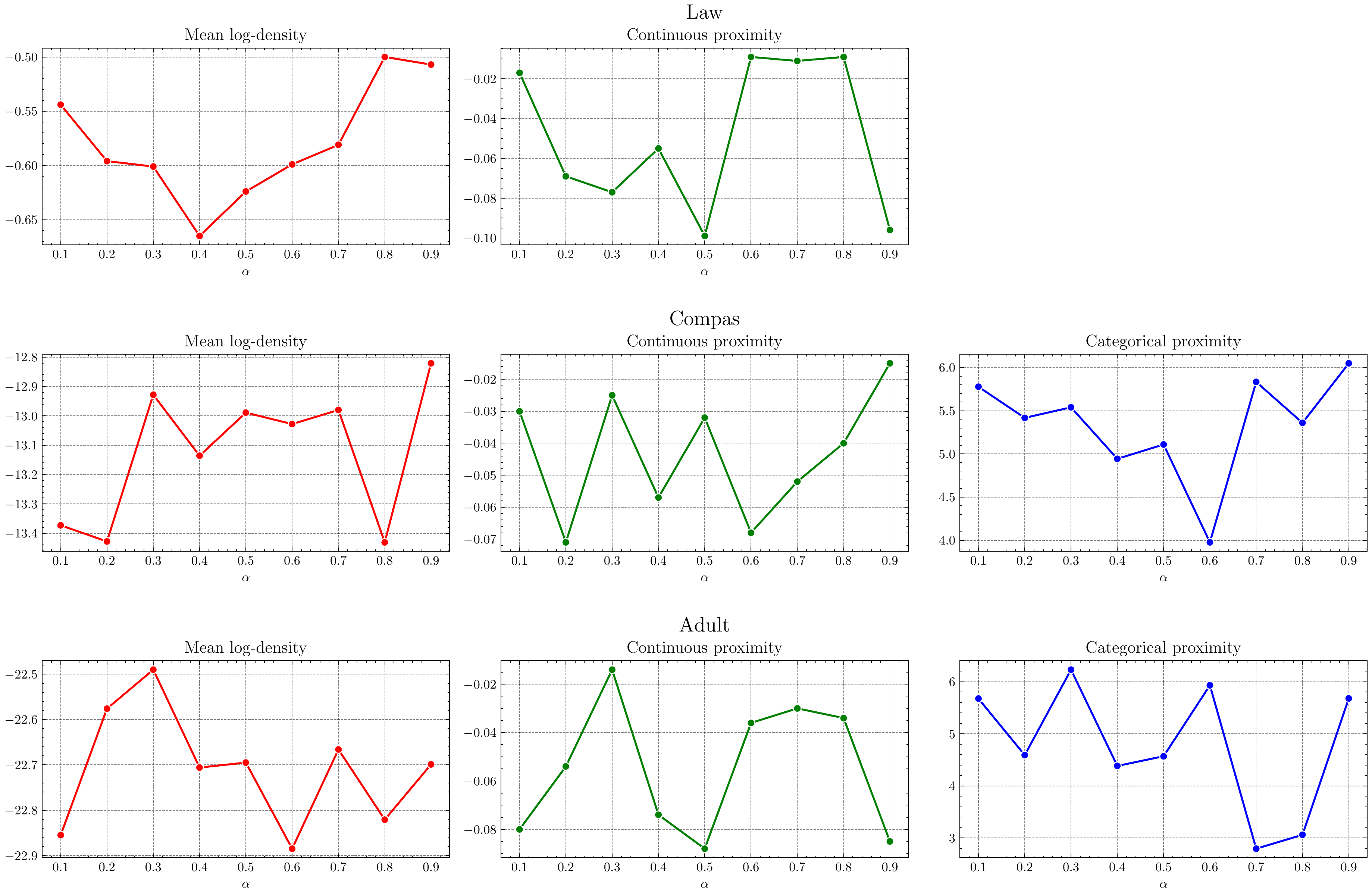}}
\caption{Our performance under different values of hyperparameter $\alpha$. Note that there are no categorical features in \texttt{Law} dataset.}
\label{fig:variation}
\end{figure}

Figure~\ref{fig:proximity} illustrates the categorical and continuous proximity. In terms of categorical proximity, our approach achieves the second-best performance with lowest variation in comparison with other approaches. The heuristic search based algorithm such as FACE and GS demonstrate the best performance in terms of this metric. Meanwhile, DiCE produces the best performance for continuous proximity, whereas CeFlow is ranked second. In general, our approach (CeFlow) achieves competitive performance in terms of proximity metric and demonstrates the least variation in comparison with others. On the other hand, Figure~\ref{fig:variation} shows the variation of our method's performance with the different values of $\alpha$. We observed that the optimal values are achieved at 0.8, 0.9 and 0.3 for \texttt{Law}, \texttt{Compas} and \texttt{Adult} dataset, respectively.

\section{Conclusion}
In this paper, we introduced a robust and efficient counterfactual explanation framework called CeFlow that utilizes the capacity of normalizing flows in generating counterfactual samples. We observed that our approach produces more stable counterfactual samples and reduces counterfactual generation time significantly. The better performance witnessed is likely because that normalizing flows can get the exact representation of the input instance and also produce the counterfactual samples by using the inverse function. Numerous extensions to the current work can be investigated upon successful expansion of normalizing flow models in interpretable machine learning in general and counterfactual explanation in specific. One potential direction is to design a normalizing flow architecture to achieve counterfactual fairness in machine learning models.

\section*{Acknowledgement}
This work is supported by the Australian Research Council (ARC) under Grant No. DP220103717, LE220100078, LP170100891, DP200101374. 


\bibliography{sample}
\bibliographystyle{splncs04}

\end{document}